\pdfoutput=1

\documentclass[11pt]{article}

\usepackage{acl}

\usepackage{times}
\usepackage{latexsym}

\usepackage[T1]{fontenc}

\usepackage[utf8]{inputenc}

\usepackage{microtype}
\usepackage{xcolor}
\usepackage{inconsolata}

\usepackage{graphicx}
\usepackage{amsmath}
\usepackage{adjustbox}
\usepackage{float}
\usepackage{multirow}
\usepackage{subfigure}
\usepackage[normalem]{ulem}
\useunder{\uline}{\ul}{}
\usepackage{listings}
\usepackage{xcolor}

\lstdefinelanguage{json}{
    morestring=[b]", %
    morestring=[b]'  %
}

\newenvironment{packed_itemize}{
\begin{list}{\labelitemi}{\leftmargin=1.5em}
  \setlength{\itemsep}{1pt}
  \setlength{\parskip}{0pt}
  \setlength{\parsep}{0pt}
  \setlength{\headsep}{0pt}
  \setlength{\topskip}{0pt}
  \setlength{\topmargin}{0pt}
  \setlength{\topsep}{0pt}
  \setlength{\partopsep}{0pt}
}{\end{list}}

\title{	Large language model as user daily behavior data generator: balancing population diversity and individual personality}

\author{
Haoxin~Li\thanks{These two authors contributed equally.},
Jingtao~Ding$^*$,
Jiahui~Gong,
Yong~Li
\\
Department of Electronic Engineering, Tsinghua University\\
dingjt15@tsinghua.org.cn; liyong07@tsinghua.edu.cn
}
\begin{document}
\maketitle
\begin{abstract}
Predicting human daily behavior is challenging due to the complexity of routine patterns and short-term fluctuations. While data-driven models have improved behavior prediction by leveraging empirical data from various platforms and devices, the reliance on sensitive, large-scale user data raises privacy concerns and limits data availability. Synthetic data generation has emerged as a promising solution, though existing methods are often limited to specific applications. In this work, we introduce BehaviorGen, a framework that uses large language models (LLMs) to generate high-quality synthetic behavior data. By simulating user behavior based on profiles and real events, BehaviorGen supports data augmentation and replacement in behavior prediction models. We evaluate its performance in scenarios such as pertaining augmentation, fine-tuning replacement, and fine-tuning augmentation, achieving significant improvements in human mobility and smartphone usage predictions, with gains of up to 18.9\%. Our results demonstrate the potential of BehaviorGen to enhance user behavior modeling through flexible and privacy-preserving synthetic data generation.
\end{abstract}

\section{Introduction}
Predicting human behavior is inherently challenging~\cite{nadkarni2016superforecasting}. While most individuals follow routine patterns shaped by behavioral habits, they also exhibit short-term fluctuations driven by specific contexts. The increasing availability of empirical data capturing user behaviors across various web platforms and smart devices has significantly enhanced our ability to forecast future behavior through the use of data-driven models~\cite{zhang2018mobility,zhang2019deep,li2022smartphone}. This progress represents a crucial step toward developing intelligent, adaptable assistants capable of supporting daily human activities~\cite{chung2018intelligent,tulshan2019survey,savcisens2023using}.

However, user behavior modeling methods rely heavily on empirical data that records real-world human behavior, often containing highly sensitive personal information. Although techniques such as privacy-preserving data publishing and privacy-aware computing approaches, like federated learning, exist, it is becoming increasingly difficult to collect large-scale, high-quality user behavior data. This limitation hinders the development of downstream applications. Synthetic data generation has emerged as a promising solution to address this data-centric challenge. Recently, deep generative models have been applied to behavior data generation in areas such as recommender systems~\cite{shi2019virtual,liu2022privacy,luo2022mindsim}, human mobility~\cite{yuan2025learning}, and urban daily life~\cite{yuan2024generating}. However, these approaches are typically tailored to specific applications and lack generalization capabilities across diverse scenarios.

Fortunately, large language models (LLMs)~\cite{zhao2023survey,GPT3}, trained on massive corpora containing extensive knowledge and capable of generating high-quality textual data, have emerged as a potential solution to this problem~\cite{long2024llms}. Recent studies have begun exploring LLMs' ability to simulate human behaviors in daily life~\cite{shao2024beyond,wang2024large,li2024more}, such as commuting and entertainment, indicating that these models have already captured significant behavioral knowledge through pretraining.

In this work, we explore the synthetic data generation capabilities of large language models (LLMs) for user behavior modeling. To meet practical requirements, modern user behavior prediction models have evolved into a two-stage paradigm~\cite{gongKDD24}. These models are first pretrained on population-level behavior data and then fine-tuned on individual-level data. Unlike textual data, both population-level diversity and individual-level specificity are crucial in determining the quality of synthetic behavior data. To address this, we introduce the BehaviorGen framework, which prompts LLMs to simulate a specific user's behavior based on a provided profile and a few real behavior events. This approach enables the flexible generation of high-quality synthetic user behavior data.

We evaluate BehaviorGen’s data generation capabilities across various usage scenarios, including:
1) \textit{Data augmentation for the pretraining stage}: where behavior diversity is critical for pretraining a generalist behavior prediction model that serves as a robust initialization for all users.
2) \textit{Data replacement for the fine-tuning stage}: where LLMs generate behaviors highly personalized to specific users, effectively replacing real data.
3) \textit{Data augmentation for the fine-tuning stage}: where, given a limited number of real records for a specific user, the model generates high-quality personalized data to supplement the fine-tuning process.
Surprisingly, we find that BehaviorGen enables LLMs to generate user behaviors that reflect both population diversity and individual personality. In the pretraining augmentation scenario, BehaviorGen achieves performance gains of up to 2.6\% and 6.9\%  in two applications: human mobility behavior prediction and smartphone usage behavior prediction, respectively. In the fine-tuning replacement scenario, synthetic data generated by BehaviorGen can replace real data, providing about 62.0\% and 87.8\% of the fine-tuning performance gains in these two applications. Finally, for the fine-tuning augmentation scenario, our results show that BehaviorGen can generate augmented data using only around 100 individual records, significantly boosting prediction performance by up to 18.9\% and 5.3\%, respectively.

\section{Related Work}

\subsection{Synthetic Data Generation with LLMs}
Synthetic data generation has gained significant momentum with the advent of large language models (LLMs) \cite{guo2024generative}. The data generated by LLMs closely approximates real-world data, making this approach a powerful solution to addressing the challenges of resource scarcity.

Designing an informative prompt is key to effective data generation with LLMs. \citet{yu2023large} explore synthetic data generation using diversely attributed prompts, which have the potential to produce diverse and richly attributed synthetic data. \citet{reynolds2021prompt} propose MetaPrompt, a method where an expanded prompt is first generated by ChatGPT, then used to further prompt LLMs for data generation. Another promising approach for task-specific data generation is to aggregate a few-shot dataset and perform parameter-efficient adaptation on the LLM \cite{guo2022improving}. \citet{chen2023mixture} train a set of soft prompt embeddings on few-shot, task-specific training data to condition the LLM for more effective text generation. \citet{he2023annollm} AnnoLLM, an LLM-powered annotation system. It first prompts LLMs to explain the reasoning behind a ground truth label, then uses these explanations to create a few-shot chain-of-thought prompt for annotating unlabeled data.

However, existing work has not adequately addressed the balance between population diversity and individual preference, a crucial consideration in user behavior generation.

\subsection{Synthetic Data for User Behavior Modeling}

Due to user privacy concerns and the difficulty of data collection, it is difficult to collect a large amount of data for model training in some user behavior domains. synthetic data generation provides a promising way. 

\citet{park2023generative} instantiate generative agents to populate an interactive sandbox environment inspired by The Sims, where end users can guide the generation of behaviors of agents using natural language. 
\citet{Zherdeva2021Prediction} use the generated synthetic data to train the Mask R-CNN framework, which is used for digital human interaction with the 3D environment. 
\citet{liu2022privacy} present UPC-SDG, a User Privacy Controllable Synthetic Data, which generates synthetic interaction data for users based on their privacy preferences to improve the performance of recommendations.
\citet{chen2021autodebias} leverage a small set of uniform synthetic data to optimize the debiasing parameters by solving the bi-level optimization problem in recommendations.
\citet{provalov2021sysevarec} propose a novel method for evaluating and comparing recommender systems using synthetic user and item data and parametric synthetic user-item response functions.

However, current work focuses on a specific domain of user behavior and lacks work on generating user behavior in all scenarios and around the clock.

\section{Preliminary}

\subsection{Behavior Data Generation Problem}
Now, we give a formal definition of our research problem:

\textsc{Problem} (User behavior generation). The user behavior can be represented as 
\[
x_i = \left(d_i, t_i, l_i, b_i, p_i \right)
\]
\textcolor{black}{where $b_i$ denotes a specific behavior occurring at location $l_i$ during time slot $t_i$ on day $d_i$. Here, $d_i$, $t_i$, $l_i$, and $b_i$ are the weekday, time slot, location, and behavior IDs, respectively. We denote the sets of weekdays, time slots, locations, and behaviors as $\mathcal{D}$, $\mathcal{T}$, $\mathcal{L}$, and $\mathcal{B}$, with sizes $N_D$, $N_T$, $N_L$, and $N_B$.}

\textcolor{black}{Additionally, $p_i$ represents the user profile, which consists of five key attributes:  \textbf{Age}: The age group of the user.\textbf{Education}: The highest education level attained by the user.\textbf{Gender}: The gender identity of the user. \textbf{Consumption}: The user's estimated consumption level.\textbf{Occupation}: The profession or job category of the user.}

User behavior sequences can be represented as 
\[
\boldsymbol{x_i} = [x_1, x_2, x_3, ..., x_I]
\]
where $I$ denotes the length of the input sequence. Our goal is to generate the user behavior sequence, which can be formulated as:
\begin{equation}
    [\hat{x_1}, \hat{x_2}, \hat{x_3}, ..., \hat{x_O}] = G([x_1, x_2, x_3, ..., x_I])
\end{equation}
where $O$ represents the length of the generated sequence and $G$ is the generation function.

\textcolor{black}{By incorporating user profiles $\mathcal{P}$ into the behavior generation process, our method ensures that the generated behavior sequences align with realistic user characteristics, leading to more accurate and personalized synthetic data.}

\subsection{Behavior Prediction Problem}
To demonstrate the effectiveness of the generated sequence, we design the user behavior prediction experiment. User behavior prediction aims to forecast future user behavior based on its past $I$ event series, which can be formed as,
\begin{equation}
\hat{b}_{t} = f(x_{t-I}, x_{t-I+1}, ..., x_{t-1})    
\end{equation}

\begin{figure*}[t]
    \centering
    \includegraphics[width = 0.9\linewidth]{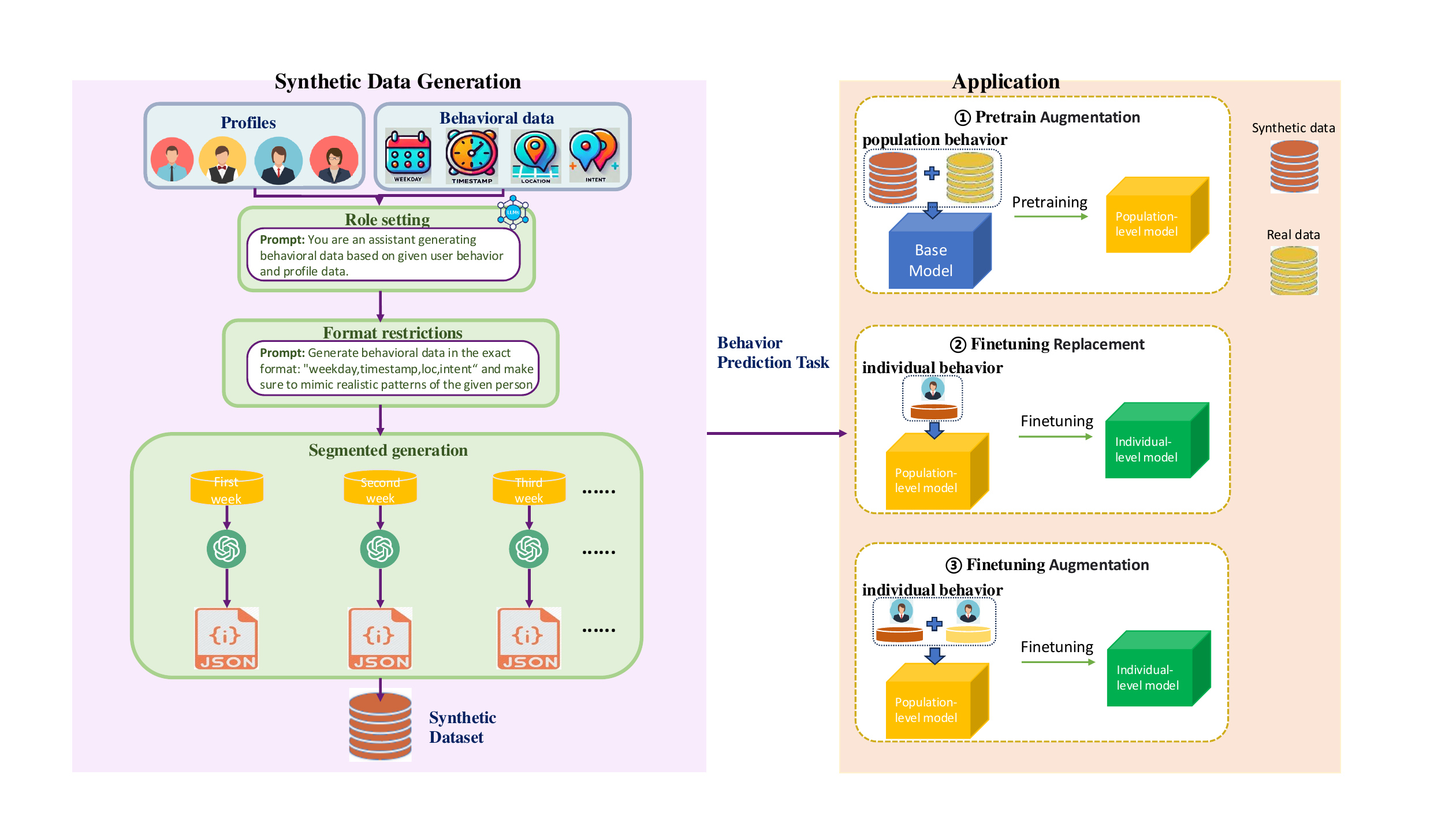}
    \vspace{-4mm}
    \caption{The Framework of BehaviorGen.}
    \vspace{-1em}
    \label{fig:Framework}
    \end{figure*}

\section{BehaviorGen Framework}
\subsection{Data Generation Procedure}

\subsubsection{Data generation Process}

\textbf{Role Setting}:  
In this stage, the Large Language Model (LLM) is assigned the role of "Generator." \textcolor{black}{We choose gpt-4o-2024-0806 model as our generator.} As shown in Figure \ref{fig:Framework}, by explicitly defining the role, the LLM is better equipped to understand the task structure and objectives, leading to more coherent and contextually appropriate output. 

\textbf{Format Restrictions}:
In order to ensure that the generated data adheres to a consistent and interpretable structure, we impose strict formatting requirements, where the output is specified as [weekday, timestamp, location, intent]. Additionally, we limit the value and scope of the generated data, such as restricting the weekday to the range of 0-6. Format restrictions ensure the validity of generated data, reducing the subsequent steps in data processing.

\textbf{Segmented Generation}:  
Given the complexity of generating long sequences of behavioral data, we utilize a segmented approach, where the user's behavior is divided into weekly segments. This reduces the risk of context drift and helps maintain consistency throughout the generation process. The segmentation is particularly useful for maintaining diversity and accuracy across varying time periods, thereby enhancing the success rate and effectiveness of data synthesis. Thus, segmentation balances diversity and faithfulness while ensuring robustness across weeks.\textcolor{black}{We give detailed experiments and explanations in the Appendix \ref{sec:weekly segment}. as to why we chose weekly segments.}

\subsection{Usage Scenarios}

\subsubsection{Pretraining Augmentation}
In real-world scenarios, application service providers can only collect a limited amount of user data, which is insufficient to support the training of a population-level behavior prediction model. Therefore, it is necessary to synthesize additional data to enhance population diversity. This need arises from the challenges of establishing a population-level model capable of capturing common behavioral patterns. To address this, we utilize a two-pronged approach involving both population-level data and pre-trained models. Specifically, we implement the PITuning\cite{gongKDD24} model architecture with both GPT-2 and Bert4Rec serving as the backbone models.

Building upon this foundational training, we incorporate behavioral data generated by the LLM as a means of data augmentation, which can be formed as,
\begin{equation}
    \mathcal{L}_p(x, \hat{x}; \theta) =  \mathcal{L}(x; \theta) +  \mathcal{L}(\hat{x}; \theta)
\end{equation}
\textcolor{black}{where $\mathcal{L}_p$ denotes the cross-entropy classification loss in the pretraining process, with $x$ and $\hat{x}$ denoting real data and synthetic data, respectively.} This method enhances the model's predictive accuracy concerning user behavior, enabling a more robust capture of behavioral patterns, as confirmed in subsequent experiments. The augmented dataset allows the model to better generalize and predict user behavior across diverse scenarios, ultimately improving its effectiveness in real-world applications.

\subsubsection{Finetuning Replacement}
Post pre-training, the fine-tuning phase serves as a pivotal step in enhancing the personalization and accuracy of recommendation systems. However, leveraging real user behavioral data in this phase poses significant privacy and security concerns. To mitigate these risks, we propose using behavioral data generated by the LLM as a replacement for real user data during fine-tuning, which can be formed as,

\begin{equation}
    \mathcal{L}_f( \hat{x}; \theta) =  \mathcal{L}(\hat{x}; \theta)
\end{equation}
\textcolor{black}{where $\mathcal{L}_f$ denotes the cross-entropy classification loss in the finetuning process.}
This approach enables the fine-tuning of the pre-trained model while preserving user privacy. Specifically, we generate behavioral data for users who included in the finetuning phase. The synthesized data is then partitioned into training, testing, and validation sets, facilitating the fine-tuning of the population-level model. This process not only enhances personalization at the individual level but also maintains the integrity of user data.

\subsubsection{Finetuning Augmentation}
Accurate prediction of long-tail user behavior within recommendation systems poses a significant challenge due to the infrequency of such data and the inherent difficulties in its collection. In response to this challenge, we advocate for a strategy that involves the synthesis of behavioral data using a limited amount of real user behavior data as a base data.

By combining LLM-generated user behavior data with this small set of authentic user data, we aim to enrich the training dataset for fine-tuning. This hybrid approach enhances the model's capacity to predict long-tail user behaviors, ensuring that even less common patterns can be adequately represented. Consequently, this not only improves the model's overall predictive capabilities but also contributes to a more comprehensive understanding of user behavior across different demographics and usage contexts.

\section{Experiment}
\subsection{Experiment Settings}
\subsubsection{Datasets}
We evaluate the performance of our model on two large-scale real-world activity datasets.

\begin{packed_itemize}
\item \textbf{Tencent Dataset.}
The Tencent Dataset consists of anonymous user trajectory data collected from October to the end of December. The dataset includes a total of 667 users and 189,954 behavioral data entries. At the population level, we select 466 users for training, while at the individual level, we use the remaining 201 users. In this dataset, we utilize location categories to represent user activities and intents.
\begin{table*}[h]
\centering
\resizebox{\textwidth}{!}{%
\begin{tabular}{cc|cccccccc|cccccccc}
\hline
\multicolumn{2}{c|}{Category} &
  \multicolumn{8}{c|}{Tencent Dataset} &
  \multicolumn{8}{c}{Smartphone Dataset} \\ \hline
\multicolumn{1}{c|}{} &
  \multirow{2}{*}{Backbone} &
  \multicolumn{4}{c|}{Bert4Rec} &
  \multicolumn{4}{c|}{PITuning} &
  \multicolumn{4}{c|}{Bert4Rec} &
  \multicolumn{4}{c}{PITuning} \\ \cline{3-18} 
\multicolumn{1}{c|}{} &
   &
  Pre &
  Rec &
  N@3 &
  \multicolumn{1}{c|}{N@5} &
  Pre &
  Rec &
  N@3 &
  N@5 &
  Pre &
  Rec &
  N@3 &
  \multicolumn{1}{c|}{N@5} &
  Pre &
  Rec &
  N@3 &
  N@5 \\ \hline
\multicolumn{1}{c|}{Real Data} &
  Pretrained &
  0.427 &
  0.466 &
  0.666 &
  \multicolumn{1}{c|}{0.663} &
  {\ul 0.418} &
  {\ul 0.449} &
  \textbf{0.667} &
  \textbf{0.661} &
  0.149 &
  0.280 &
  0.515 &
  \multicolumn{1}{c|}{0.551} &
  0.123 &
  0.168 &
  0.435 &
  0.468 \\ \hline
\multicolumn{1}{c|}{\multirow{5}{*}{\begin{tabular}[c]{@{}c@{}}Real Data +\\ Synthetic Data\end{tabular}}} &
  SeqGAN &
  0.417 &
  0.452 &
  0.682 &
  \multicolumn{1}{c|}{0.676} &
  0.401 &
  0.429 &
  0.638 &
  0.630 &
  0.150 &
  0.259 &
  0.524 &
  \multicolumn{1}{c|}{0.554} &
  0.134 &
  0.162 &
  0.435 &
  0.463 \\
\multicolumn{1}{c|}{} &
  DiffuSeq &
  {\ul 0.436} &
  {\ul 0.471} &
  {\ul 0.684} &
  \multicolumn{1}{c|}{{\ul 0.685}} &
  0.281 &
  0.366 &
  0.624 &
  0.620 &
  0.167 &
  0.283 &
  0.524 &
  \multicolumn{1}{c|}{0.557} &
  {\ul 0.136} &
  {\ul 0.174} &
  0.433 &
  0.467 \\
\multicolumn{1}{c|}{} &
  UPC\_SDG &
  0.424 &
  0.457 &
  0.676 &
  \multicolumn{1}{c|}{0.670} &
  0.384 &
  0.417 &
  0.630 &
  0.632 &
  {\ul 0.188} &
  {\ul 0.295} &
  {\ul 0.528} &
  \multicolumn{1}{c|}{{\ul 0.558}} &
  0.130 &
  0.169 &
  {\ul 0.438} &
  {\ul 0.472} \\
\multicolumn{1}{c|}{} &
  Ours &
  \textbf{0.447} &
  \textbf{0.480} &
  \textbf{0.702} &
  \multicolumn{1}{c|}{\textbf{0.694}} &
  \textbf{0.426} &
  \textbf{0.450} &
  {\ul 0.655} &
  {\ul 0.659} &
  \textbf{0.213} &
  \textbf{0.315} &
  \textbf{0.543} &
  \multicolumn{1}{c|}{\textbf{0.570}} &
  \textbf{0.201} &
  \textbf{0.186} &
  \textbf{0.454} &
  \textbf{0.479} \\ \cline{2-18} 
\multicolumn{1}{c|}{} &
  Improvement &
  2.5\% &
  1.9\% &
  2.6\% &
  \multicolumn{1}{c|}{1.3\%} &
  1.9\% &
  0.2\% &
  -1.8\% &
  -0.3\% &
  13.3\% &
  6.8\% &
  2.8\% &
  \multicolumn{1}{c|}{2.2\%} &
  4.8\% &
  6.9\% &
  3.7\% &
  1.5\% \\ \hline
\end{tabular}
}
\caption{Overall prediction performance Pretrain Augmentation compared with baselines on Tencent and Smartphone datasets. The improvement here is calculated using the formula: (ours - the best result from pretrained and baseline) / the best result from pretrained and baseline.}
\label{tab:pretrain augmentation}
\end{table*}
\item \textbf{Smartphone Dataset.} 
The Smartphone Dataset is sampled from the usage log of the mobile phones. When a user uses mobile phones, various types of logs are generated, desensitized and reported (with user consent). We selected 114 types of events that are commonly monitored in most mobile applications and classified them into 18 intents, which cover the aspects of news, study, work, entertainment, sports, etc. We sampled two datasets between June 1st and August 22nd, 2023 (the first) and August 22nd and September 10th, 2023 (the second) which in total contain 4,500 and 5,000 anonymous users.
\end{packed_itemize}

\subsubsection{Metrics}
To assess model performance, we employ four widely used metrics: precision ($Pre$), recall ($Rec$), and NDCG(N)~\cite{ding2020improving,ding2020simplify}. NDCG gauge classification accuracy and ranking quality, respectively, while Pre and Rec evaluate the average prediction accuracy for each intent, indicating the model's predictive quality across intents.  Refer to Appendix \ref{metrics} for metric calculations.

\subsubsection{Baselines}
We carefully select the following three representative methods to compare with our proposed algorithm, which include generative methods for sequence data (SeqGAN~\cite{yu2017seqgan}), diffusion-based sequence generation models (DiffuSeq~\cite{gong2022diffuseq}), and a synthetic data generation method (UPC\_SDG~\cite{liu2022privacy}). We provide the details of baselines in Appendix \ref{baseline}.

\subsubsection{Evaluation Backbones.}
We choose PITuning~\cite{gongKDD24} and Bert4Rec~\cite{sun2019bert4rec} as the evaluation backbone.

$\bullet$ \textbf{PITuning} PITuning is a Population-Individual Tuning framework that enhances common pattern extraction through dynamic event-to-intent transition modeling and addresses long-tailed preferences via adaptive unlearning strategies.

$\bullet$ \textbf{Bert4Rec} Bert4Rec, a bidirectional encoder representation from Transformers, enhances the power of the historical sequence representations by jointly conditioning the left and right context.

\subsubsection{Tasks}
$\bullet$ \textbf{Pretraining Augmentation} In the Pretraining phase, as shown in Figure \ref{fig:Framework}, we leverage population-level data in combination with synthetic data generated by our framework using large language models (LLMs). This synthetic data is employed as a form of data augmentation.

$\bullet$ \textbf{Finetuning Replacement} In the fine-tuning phase, as shown in Figure \ref{fig:Framework}, we synthesized a set of personalized data from individual-level user data to protect user privacy. This synthetic data was used to replace actual user personal data. We fine-tuned a population-level model, which had been pre-trained, using this synthetic dataset.

$\bullet$ \textbf{Finetuning Augmentation} During the finetuning phase, a common challenge arises when dealing with new users who possess limited behavioral data, which may be insufficient for effective model adaptation. To address this issue, we propose a method for generating personalized behavioral data based on the limited real individual data(about 105 logs) available from users. This synthesized data serves to enhance the finetuning results of individual models.

\begin{table*}[h]
\centering
\resizebox{\textwidth}{!}{%
\begin{tabular}{cc|llllllll|cccccccc}
\hline
\multicolumn{2}{c|}{Category} &
  \multicolumn{8}{c|}{Tencent Dataset} &
  \multicolumn{8}{c}{Smartphone Dataset} \\ \hline
\multicolumn{1}{c|}{} &
  \multirow{2}{*}{Backbone} &
  \multicolumn{4}{c|}{Bert4Rec} &
  \multicolumn{4}{c|}{PITuning} &
  \multicolumn{4}{c|}{Bert4Rec} &
  \multicolumn{4}{c}{PITuning} \\ \cline{3-18} 
\multicolumn{1}{c|}{} &
   &
  \multicolumn{1}{c}{Pre} &
  \multicolumn{1}{c}{Rec} &
  \multicolumn{1}{c}{N@3} &
  \multicolumn{1}{c|}{N@5} &
  \multicolumn{1}{c}{Pre} &
  \multicolumn{1}{c}{Rec} &
  \multicolumn{1}{c}{N@3} &
  \multicolumn{1}{c|}{N@5} &
  Pre &
  Rec &
  N@3 &
  \multicolumn{1}{c|}{N@5} &
  Pre &
  Rec &
  N@3 &
  N@5 \\ \hline
\multicolumn{1}{c|}{\multirow{2}{*}{Real Data}} &
  Pretrained &
  0.447 &
  0.474 &
  0.693 &
  \multicolumn{1}{l|}{0.691} &
  0.422 &
  0.454 &
  0.684 &
  0.678 &
  0.207 &
  0.340 &
  0.542 &
  \multicolumn{1}{c|}{0.568} &
  0.126 &
  0.178 &
  0.440 &
  0.478 \\
\multicolumn{1}{c|}{} &
  Finetuned &
  0.597 &
  0.614 &
  0.790 &
  \multicolumn{1}{l|}{0.791} &
  0.583 &
  0.604 &
  0.780 &
  0.774 &
  0.322 &
  0.366 &
  0.594 &
  \multicolumn{1}{c|}{0.614} &
  0.306 &
  0.355 &
  0.627 &
  0.668 \\ \hline
\multicolumn{1}{c|}{\multirow{5}{*}{Synthetic Data}} &
  SeqGAN &
  0.185 &
  0.221 &
  0.381 &
  \multicolumn{1}{l|}{0.375} &
  0.194 &
  0.228 &
  0.394 &
  0.392 &
  0.288 &
  0.309 &
  0.542 &
  \multicolumn{1}{c|}{0.577} &
  0.227 &
  0.296 &
  0.576 &
  0.616 \\
\multicolumn{1}{c|}{} &
  DiffuSeq &
  \multicolumn{1}{c}{0.152} &
  \multicolumn{1}{c}{0.223} &
  \multicolumn{1}{c}{0.409} &
  \multicolumn{1}{c|}{0.409} &
  \multicolumn{1}{c}{0.161} &
  \multicolumn{1}{c}{0.234} &
  \multicolumn{1}{c}{0.417} &
  \multicolumn{1}{c|}{0.425} &
  0.233 &
  \textbf{0.340} &
  0.550 &
  \multicolumn{1}{c|}{0.589} &
  0.228 &
  0.301 &
  0.591 &
  0.628 \\
\multicolumn{1}{c|}{} &
  UPC\_SDG &
  \multicolumn{1}{c}{0.172} &
  \multicolumn{1}{c}{0.148} &
  \multicolumn{1}{c}{0.229} &
  \multicolumn{1}{c|}{0.223} &
  \multicolumn{1}{c}{0.170} &
  \multicolumn{1}{c}{0.159} &
  \multicolumn{1}{c}{0.236} &
  \multicolumn{1}{c|}{0.234} &
  0.280 &
  0.315 &
  0.543 &
  \multicolumn{1}{c|}{0.569} &
  0.260 &
  0.317 &
  0.562 &
  0.585 \\
\multicolumn{1}{c|}{} &
  Ours &
  \textbf{0.540} &
  \textbf{0.539} &
  \textbf{0.746} &
  \multicolumn{1}{l|}{\textbf{0.734}} &
  \textbf{0.516} &
  \textbf{0.529} &
  \textbf{0.733} &
  \textbf{0.724} &
  \textbf{0.308} &
  0.334 &
  \textbf{0.568} &
  \multicolumn{1}{c|}{\textbf{0.593}} &
  \textbf{0.270} &
  \textbf{0.333} &
  \textbf{0.602} &
  \textbf{0.643} \\ \cline{2-18} 
\multicolumn{1}{c|}{} &
  Replacement &
  62.0\% &
  46.4\% &
  54.4\% &
  \multicolumn{1}{l|}{43.0\%} &
  58.4\% &
  50.0\% &
  51.0\% &
  47.9\% &
  87.8\% &
  -23.1\% &
  50.0\% &
  \multicolumn{1}{c|}{54.3\%} &
  80.0\% &
  87.6\% &
  86.6\% &
  86.8\% \\ \hline
\end{tabular}
}
\caption{Overall prediction performance Finetuning Replacement compared with baselines on Tencent and Smartphone datasets. The replacement here is calculated using the formula: (ours - pretrained) / (real data finetuned - pretrained). \textcolor{black}{Replacement tells us what percent of the improvement we can achieve when fine-tuning the model using only our synthetic data compared to using the real data.}}
\vspace{-1em}
\label{tab:Finetuning Replacement}
\end{table*}

\begin{table*}[ht]
\centering
\resizebox{\textwidth}{!}{%
\begin{tabular}{cc|cccccccc|cccccccc}
\hline
\multicolumn{2}{c|}{Category} &
  \multicolumn{8}{c|}{Tencent Dataset} &
  \multicolumn{8}{c}{Smartphone Dataset} \\ \hline
\multicolumn{1}{c|}{} &
  \multirow{2}{*}{Backbone} &
  \multicolumn{4}{c|}{Bert4Rec} &
  \multicolumn{4}{c|}{PITuning} &
  \multicolumn{4}{c|}{Bert4Rec} &
  \multicolumn{4}{c}{PITuning} \\ \cline{3-18} 
\multicolumn{1}{c|}{} &
   &
  Pre &
  Rec &
  N@3 &
  \multicolumn{1}{c|}{N@5} &
  Pre &
  Rec &
  N@3 &
  N@5 &
  Pre &
  Rec &
  N@3 &
  \multicolumn{1}{c|}{N@5} &
  Pre &
  Rec &
  N@3 &
  N@5 \\ \hline
\multicolumn{1}{c|}{Limited Real Data} &
  Finetuned &
  {\ul 0.495} &
  {\ul 0.528} &
  \textbf{0.709} &
  \multicolumn{1}{c|}{\textbf{0.715}} &
  {\ul 0.455} &
  {\ul 0.493} &
  {\ul 0.697} &
  {\ul 0.695} &
  0.322 &
  0.366 &
  0.594 &
  \multicolumn{1}{c|}{0.614} &
  0.306 &
  0.355 &
  0.627 &
  0.668 \\ \hline
\multicolumn{1}{c|}{\multirow{5}{*}{\begin{tabular}[c]{@{}c@{}}Limited Real Data +\\ Synthetic Data\end{tabular}}} &
  SeqGAN &
  0.261 &
  0.297 &
  0.494 &
  \multicolumn{1}{c|}{0.496} &
  0.251 &
  0.287 &
  0.488 &
  0.493 &
  0.331 &
  0.377 &
  {\ul 0.600} &
  \multicolumn{1}{c|}{{\ul 0.621}} &
  0.315 &
  0.343 &
  0.624 &
  {\ul 0.675} \\
\multicolumn{1}{c|}{} &
  DiffuSeq &
  0.219 &
  0.298 &
  0.494 &
  \multicolumn{1}{c|}{0.501} &
  0.207 &
  0.282 &
  0.504 &
  0.503 &
  0.333 &
  0.376 &
  0.596 &
  \multicolumn{1}{c|}{{\ul 0.621}} &
  {\ul 0.316} &
  {\ul 0.356} &
  0.628 &
  0.674 \\
\multicolumn{1}{c|}{} &
  UPC\_SDG &
  0.309 &
  0.277 &
  0.407 &
  \multicolumn{1}{c|}{0.419} &
  0.277 &
  0.278 &
  0.435 &
  0.439 &
  {\ul 0.339} &
  {\ul 0.378} &
  {\ul 0.600} &
  \multicolumn{1}{c|}{{\ul 0.621}} &
  0.308 &
  0.354 &
  {\ul 0.635} &
  0.672 \\
\multicolumn{1}{c|}{} &
  Ours &
  \textbf{0.545} &
  \textbf{0.547} &
  {\ul 0.708} &
  \multicolumn{1}{c|}{{\ul 0.709}} &
  \textbf{0.541} &
  \textbf{0.541} &
  \textbf{0.728} &
  \textbf{0.722} &
  \textbf{0.345} &
  \textbf{0.398} &
  \textbf{0.612} &
  \multicolumn{1}{c|}{\textbf{0.635}} &
  \textbf{0.328} &
  \textbf{0.364} &
  \textbf{0.643} &
  \textbf{0.682} \\ \cline{2-18} 
\multicolumn{1}{c|}{} &
  Improvement &
  10.1\% &
  3.5\% &
  -0.1\% &
  \multicolumn{1}{c|}{-0.8\%} &
  18.9\% &
  9.7\% &
  4.4\% &
  3.9\% &
  1.8\% &
  5.3\% &
  2\% &
  \multicolumn{1}{c|}{2.3\%} &
  3.8\% &
  2.2\% &
  1.3\% &
  1.0\% \\ \hline
\end{tabular}
}
\caption{Overall prediction performance Finetuning Augmentation compared with baselines on Tencent and Smartphone datasets. The improvement here is calculated using the formula: (ours - the best result from pretrained and baseline) / the best result from pretrained and baseline.}
\vspace{-1em}
\label{tab:Finetuning Augmentation}
\end{table*}

\subsection{Overall Performance Analysis}
We report experiments on three usage scenarios for two prediction applications (Table~\ref{tab:pretrain augmentation}-\ref{tab:Finetuning Augmentation}).
Across all experiments, our framework demonstrates clear superiority over baseline methods, both in terms of performance metrics and its ability to produce high-quality synthetic data. Whether augmenting real data during pretraining or replacing sensitive user data during finetuning, the models trained with our synthetic data consistently outperform those trained with data generated by SeqGAN, DiffuSeq, and UPC\_SDG. The fine balance between diversity and faithfulness achieved by our framework not only leads to enhanced model performance but also offers a scalable solution for privacy-preserving data generation.

$\bullet$ \textbf{Our method demonstrates minimal discrepancies compared to fine-tuning with real data.} Specifically, the proposed framework effectively generates personalized synthetic data, crucial for maintaining performance levels that closely resemble those achieved through fine-tuning on real data, all while ensuring user privacy. As evidenced in Table \ref{tab:Finetuning Replacement}, models fine-tuned using synthetic data exhibit a performance gap of merely 5.7\% and 1.4\% in average precision, achieving scores of 0.540 on the Tencent dataset and 0.308 on the Smartphone dataset, respectively. Furthermore, the average replacement rate of 57.6\% highlights the equilibrium our framework achieves between privacy preservation and model efficacy. Notably, our synthetic data generation method surpasses other techniques, including SeqGAN and UPC\_SDG, by a considerable margin, thereby affirming our framework’s efficacy in accurately capturing individual-level user behavior patterns.

$\bullet$ \textbf{Population-Level Analysis}
In the pre-training phase, as shown in Figure \ref{fig:Framework}, we performed data augmentation using population-level data combined with synthetic data. The emphasis during this phase was on extracting common features across the population. The introduction of synthetic data not only enriched the diversity of user behavior patterns but also maintained a high level of faithfulness to real user data. This balance between diversity and fidelity enabled the model to generalize more effectively. As presented in Table \ref{tab:pretrain augmentation}, models pre-trained with a mix of real and synthetic data exhibited significant improvements in accuracy and recall, indicating that synthetic data introduces sufficient variability without compromising the coherence of user trajectories.

$\bullet$ \textbf{Individual-Level Analysis}
In the fine-tuning phase, we synthesized a personalized dataset derived from individual user data to replace real user data. This approach not only ensures privacy but also faithfully captures individualized behavior patterns critical for intent prediction and behavior modeling tasks. At the individual level, our synthetic data remains faithful to real user behaviors while introducing subtle variations that better capture users' distinct decision-making processes.

As shown in Tables \ref{tab:Finetuning Replacement} and \ref{tab:Finetuning Augmentation}, models fine-tuned with synthetic data significantly outperformed those fine-tuned solely on real data, particularly in metrics such as NDCG@3 and NDCG@5. The higher NDCG scores highlight that synthetic data more effectively mirrors individual users' preferences, improving the model's performance in recommendation tasks. Furthermore, the coherence of the generated user trajectories, a critical aspect of faithfulness, ensures that the synthetic data aligns with the unique patterns of real users, thereby enhancing both intent and event prediction accuracy.

\subsection{Ablation Study}
In this study, we conducted a comprehensive evaluation of our proposed method through a series of ablation experiments, which were designed to assess the impact of various components on the quality of the generated behavioral data. The results of these experiments are summarized in Table \ref{tab:ablation} and include several key performance indicators that are critical for evaluating the efficacy of our approach.

\begin{table}[ht]
    \centering
\resizebox{0.45\textwidth}{!}{\begin{tabular}{|c|cccc|c|}
\hline
Method      & \multicolumn{1}{c|}{KS\_P}      & BLEU         & \multicolumn{1}{c|}{BD} & JSD          & Pass@1 \\ \hline
no\_profile & \multicolumn{1}{c|}{0.231}       & 0.444          & 0.068                    & 0.053          & 100\%  \\
no\_role    & \multicolumn{1}{c|}{0.213}       & 0.449          & 0.066                    & 0.053          & 97\%   \\
our         & \multicolumn{1}{c|}{{\ul 0.327}} & \textbf{0.512} & {\ul 0.050}              & \textbf{0.029} & 100\%  \\ \hline
no\_segment    & \textbf{0.489}                   & {\ul 0.492}    & \textbf{0.035}           & {\ul 0.041}    & 22.5\% \\
no\_format  & nan                              & nan            & nan                      & nan            & 0\%    \\ \hline
\end{tabular}}
    \caption{The table presents the results comparing various methods in data generation based on several evaluation metrics: KS\_P, BLEU, BD, JSD, and Pass@1. The highest value for $KS_P$ and $BLEU$ and the lowest value for $BD$ and $JSD$ are highlighted.}
    \label{tab:ablation}
\end{table}

We use following metrics:
\textbf{KS\_P} measures the discrepancy between the distributions of generated and real data, with higher values indicating better alignment. \textbf{BLEU} assesses the n-gram overlap between generated and reference text, where a higher score signifies greater textual similarity. \textbf{BD} quantifies the similarity between two probability distributions, with lower values indicating greater similarity. \textbf{JSD} evaluates the similarity between distributions, ranging from 0 to 1, where lower scores denote closer alignment. Finally, \textbf{Pass@1} reflects the proportion of instances where the model successfully predicts user behavior. 

\textbf{Profile information}: As shown in Table \ref{tab:ablation}, profile information significantly improves model performance, with KS\_P increasing from 0.231 to 0.327 and JSD decreasing from 0.053 to 0.029, indicating better distribution alignment and enhanced generation quality.

    \textbf{Role setting}: The "no\_role" method shows moderate performance in KS\_P and BLEU, indicating that including role information positively impacts the diversity and coherence of the generated output. The relatively low BD and JSD values suggest that this method produces a more faithful representation of the target distribution. The high Pass@1 score indicates that users could successfully identify correct outputs in 97\% of cases, which is commendable.
    
    \textbf{Format restrictions}: The "no\_format" method shows NaN values across all metrics, indicating that this setting was unable to produce outputs in the correct format, resulting in a complete loss of data usability. The 0\% Pass@1 further emphasizes that the outputs were entirely unusable, underscoring the critical role of format restrictions in generating coherent and interpretable results. This implies that neglecting format considerations severely hampers the model's ability to produce valid outputs.
    
    \textbf{Segmented generation}: The "no\_segment" method achieves the highest KS\_P score and a competitive BLEU score, suggesting that segmenting the data enhances diversity and textual coherence significantly. The low BD and JSD values indicate that this method produces outputs that are closely aligned with the intended data distribution, improving the quality of the generated content. However, the low Pass@1 score (22.5\%) implies that while the outputs are diverse and coherent, they may not be entirely aligned with user expectations or specific intents, leading to a lower success rate in identifying correct outputs. Therefore, we adopt a segmented generation approach combined with Role setting and Format restrictions, ensuring the generated data maintains both diversity and fidelity while consistently producing effective and usable outputs.The prompt used in our method can be found in Appendix \ref{sec:prompt}.

\subsection{Case Study: Intent Distribution Analysis}

In this case study, we analyze the intent distribution at both the population level and individual user level to demonstrate the necessity and effectiveness of the fine-tuning phase in our model. Specifically, we examine how well the synthetic data captures individual users' intent distributions compared to the population-level distribution.

\begin{figure}[t]
    \centering
    \subfigure[User A]{
        \includegraphics[width=0.75\linewidth]{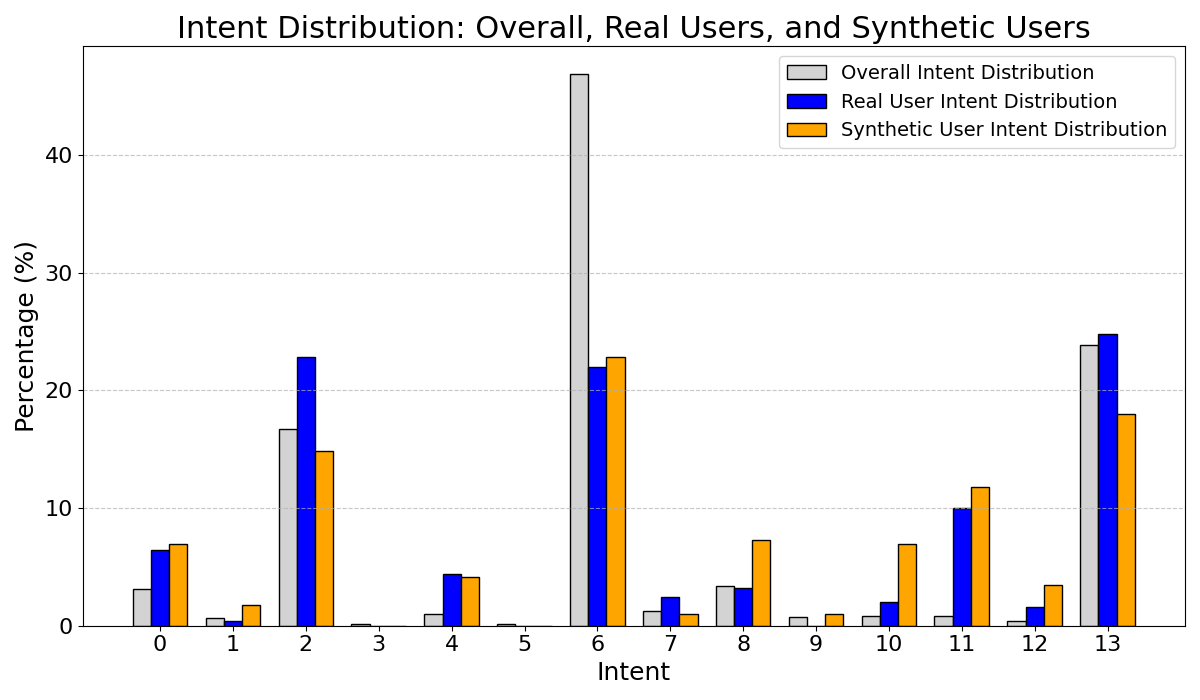} %
        \label{fig:UserIntenA}
    }
    \subfigure[User B]{
        \includegraphics[width=0.75\linewidth]{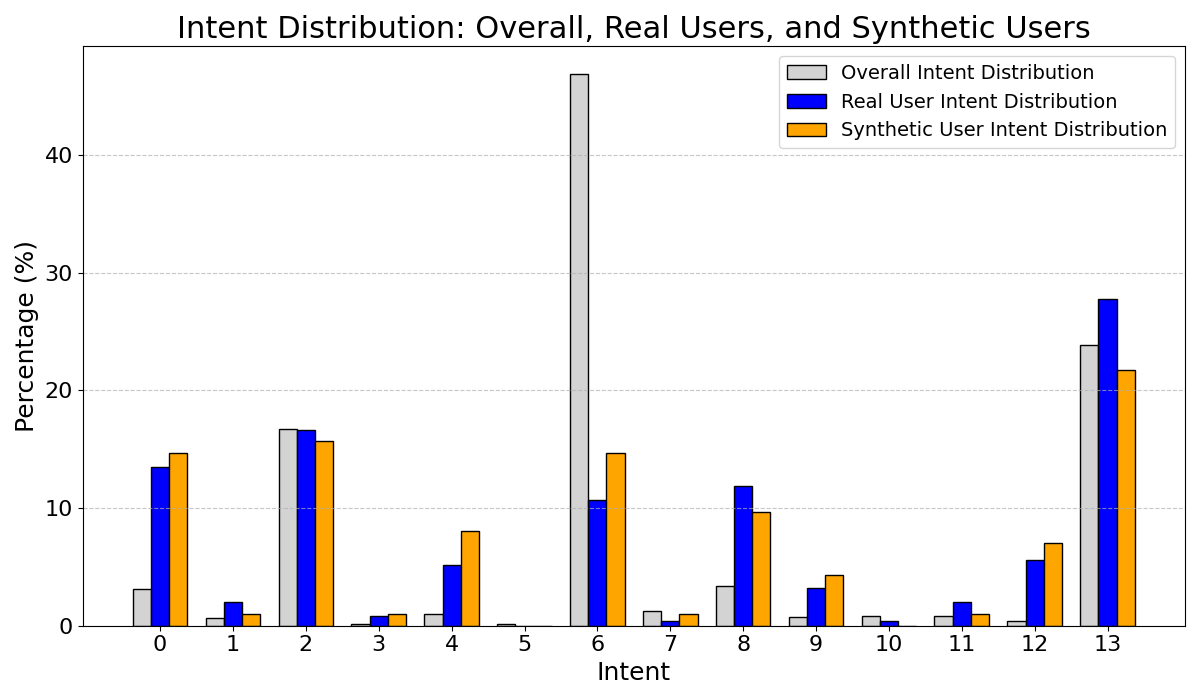} %
        \label{fig:UserIntentB}
    }
    \vspace{-1em}
    \caption{population and individual intent distribution.}
    \vspace{-1em}
    \label{fig:case study}
\end{figure}

We selected two users, User A and User B, for a comparative analysis of their intent distributions. Figure \ref{fig:UserIntenA} and Figure \ref{fig:UserIntentB} present the real intent distributions of these users alongside the intent distributions generated from synthetic data.
And the population level intent distribution is shown in grey.
For both User A and User B, the real intent distribution (shown in blue) demonstrates a pronounced deviation from the population-level distribution.  In contrast, the synthetic data (shown in orange) reflects a strong alignment with the real intent distribution, validating the hypothesis that synthetic data can faithfully represent individual user behaviors.

The discrepancies between the population-level intent distribution and the individual user intent distributions emphasize the necessity of the fine-tuning phase. By utilizing synthetic data tailored to reflect individual users' intents, we can enhance the model's performance in personalized recommendation tasks. The findings from this analysis confirm that the fidelity of synthetic data is crucial, as it ensures that the model not only generalizes well across the population but also effectively adapts to the unique preferences of individual users.

\section{Conclusion}

This preliminary study explores the potential of large language models (LLMs) for generating synthetic user behavior data. Experimental results across three synthetic data usage scenarios show promising performance in enhancing two downstream behavior prediction applications. These findings suggest that the generated synthetic behavior data effectively captures both population-level diversity and individual-level specificity, reflecting the complexity of human daily behavioral patterns.

\section{Limitations}

\textbf{Ethical Considerations.} The ethical implications of using real behavior data in this study are of utmost importance. While the data we used is anonymized and preprocessed by our providers using privacy-preserving techniques, including differential privacy, to prevent any risk of personal identification, it is still necessary to address potential concerns around privacy. The use of differential privacy ensures that individual-level data cannot be reconstructed from aggregated information, further strengthening data security. We have signed non-disclosure agreements (NDAs) with our data providers and work under their supervision to ensure responsible data handling and analysis. 

\textbf{Bias.} Since our work uses real user data to prompt LLMs in generating synthetic behavior data, there are two potential sources of bias. The first source is the empirical data provided, which may not equally represent all user groups, potentially leading to biases in how certain behaviors or demographics are modeled. The second source of bias stems from the LLMs themselves, which may exhibit biases based on the composition of their pretraining corpus, reflecting imbalances or stereotypes present in the data they were trained on. To address these concerns, we plan to implement several mitigation strategies. This includes applying fairness-aware techniques during both data preprocessing and model prompting to ensure diverse and equitable representation across user groups. 

\textbf{Future Directions.} There are several areas where our work can be further enhanced. First, developing more data-efficient generation methods is crucial, as behavior prediction scenarios typically involve large volumes of training data. Reducing the dependency on massive datasets without compromising model performance would significantly improve scalability and practicality. Second, improving the underlying LLMs to better understand and model human daily activities will be key to generating higher-quality synthetic data~\cite{ding2024understanding,zhang2025survey}.

\bibliography{custom}

\appendix
\clearpage
\section{Appendix}
\label{sec:appendix}

\subsection{Details of Metrics}
\label{metrics}
we employ four widely used metrics:precision ($Pre$), recall ($Rec$), and NDCG(N). The calculation of each metric is as follows.
The formula for $Pre$ :
\begin{equation}\label{equ:Pre}
    Pre = = \frac{1}{|C|} \sum_{c \in C} \frac{\text{TP}_c}{\text{TP}_c + \text{FP}_c}
\end{equation}
The formula for $Rec$ :
\begin{equation}\label{equ:Rec}
    Rec = \frac{1}{|C|} \sum_{c \in C} \frac{\text{TP}_c}{\text{TP}_c + \text{FN}_c}
\end{equation}
Where $|C|$ represents the total number of classes, True Positives $\text({TP}_c)$ denotes the number of samples correctly classified as class $c$, False Positives $\text({FP}_c)$ represents the number of samples incorrectly classified as class $c$, and False Negatives $\text({FN}_c)$ stands for the number of samples incorrectly classified as other classes instead of class $c$. And $\text{Precision}$ and $\text{Recall}$ respectively refer to the precision and recall of class $c$.\\
The formula for $N@k$ :
\begin{equation}\label{equ:N@K}
    N@k = \frac{\sum_{i=1}^{K} \frac{2^{rel_i} - 1}{\log_2(i+1)}}{\sum_{j=1}^{|REL_K|} \frac{rel_j - 1}{\log_2(j+1)}}
\end{equation}
where $rel_i$ means the graded relevance of the result at position $i$, and $|REL_K|$ means the list of predictions in the result ranking list up to position $K$.

\subsection{Details of Baselines}
\label{baseline}
Here we introduce the details of each baseline.

\begin{packed_itemize}

\item \textbf{SeqGAN~\cite{yu2017seqgan}.}
SeqGAN is a sequence generative adversarial network that models sequence data generation as a reinforcement learning task, utilizing a GAN structure to capture the sequential dependencies in data generation.

\item \textbf{DiffuSeq~\cite{gong2022diffuseq}.}
DiffuSeq is a diffusion-based sequence generation model that adapts the diffusion process for text and sequence data generation, offering state-of-the-art performance on various generative tasks by leveraging noise-perturbed transitions during generation.

\item \textbf{UPC\_SDG~\cite{liu2022privacy}.}
UPC\_SDG is a user trajectory synthetic data generation model, which focuses on preserving the statistical characteristics of the original data. It generates plausible user trajectories by maintaining important spatiotemporal relationships and is particularly effective for data privacy scenarios.

\end{packed_itemize}

\subsection{Used Prompts}
\label{sec:prompt}

\begin{figure}[h]
    \centering
    \begin{minipage}{0.45\textwidth} %
        \begin{lstlisting}[language=json]
messages = [
    {
        "role": "system",
        "content": """
You are an assistant generating behavioral data based on given user behavior and profile data. I will provide you with a subset of real behavioral data in the format [weekday, timestamp, loc, intent].

Your task:
1. Generate behavioral data for one month (minimum 90 lines) in the exact format: "weekday,timestamp,loc,intent".
2. Make sure to mimic realistic patterns of the given person, such as daily routines, work hours, and leisure activities, while ensuring diversity in location (loc) and intent. Don't have repetitive generation.
3. Ensure the weekdays values are within the range of 0-6, and timestamp values are within the range of 0-95.
4. Ensure that generated data has more than 100 lines and is in the correct format.
    },
    {
        "role": "user",
        "content": f"Profile:\n{json.dumps(user_profile)}\nBehavior data:\n{behavior_part.to_string(index=False)}"
    }
]
        \end{lstlisting}
        \caption{Prompt for generating behavioral data.}
        \label{fig:behavioral_data_prompt}
    \end{minipage}
\end{figure}

\subsection{\textcolor{black}{Study of segment}}
\label{sec:weekly segment}
\begin{figure}[t]
    \centering
    \includegraphics[width=\linewidth]{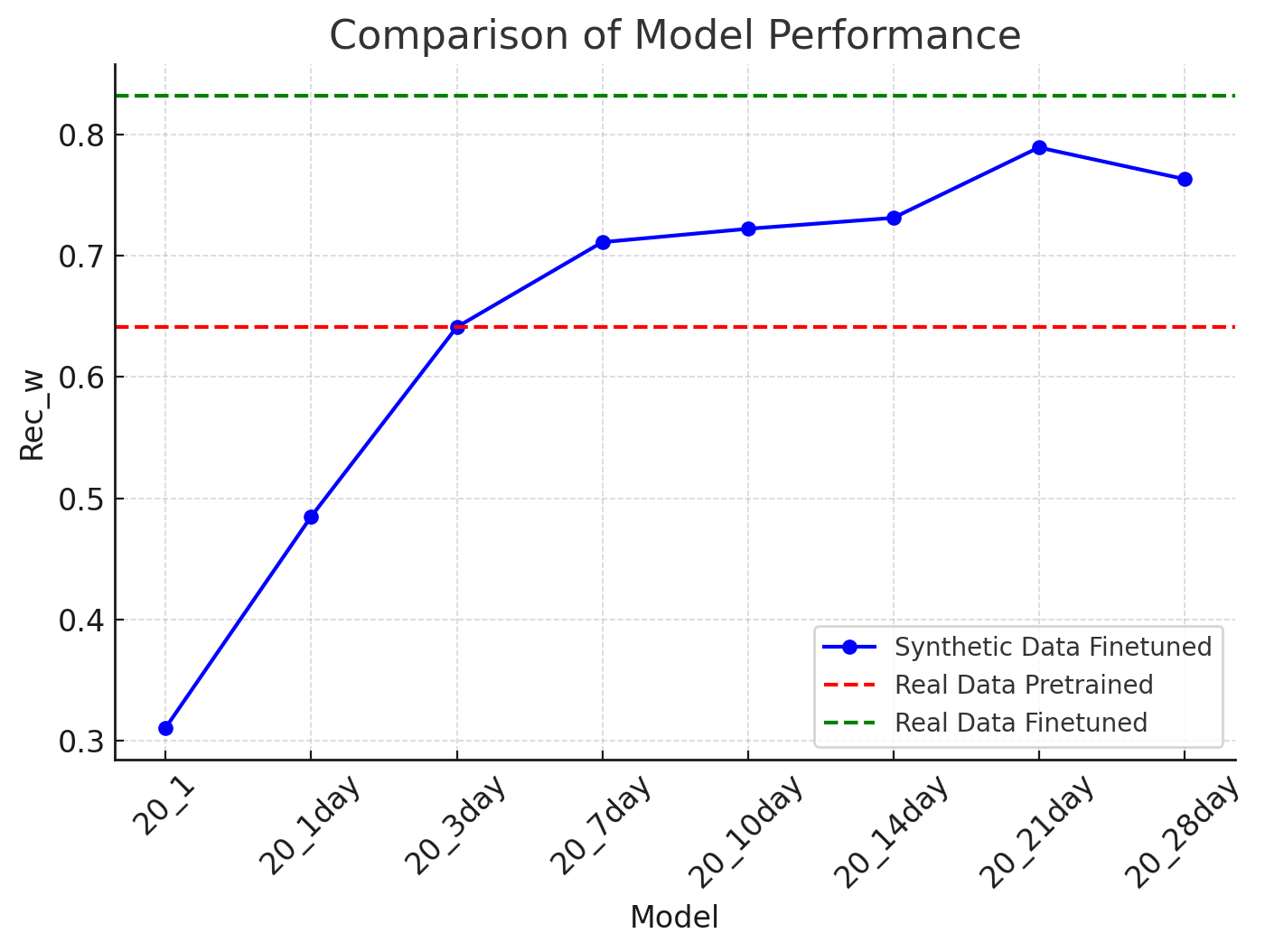} %
    \caption{segment study}
    \label{fig:segment_study}
\end{figure}

\textcolor{black}{We did experiments on segments on a small scale before generating synthetic data for all users of the dataset. We randomly select a batch of users (20), and give LLM users' 1 piece, 1 day, 3 days, 7 days, 10 days......of real data and then fine-tuned on the pre-trained model with the generated synthetic data to see how the metrics change, as shown in Figure\ref{fig:segment_study}.It can be seen that when 7 days of data are provided to LLM, the effect of synthetic data is close to convergence.The line charts of the other metrics except Rec also show this trend. Although more data is provided, weekly segment is considered as the best choice for cost and benefit considerations.}

\subsection{\textcolor{black}{Study of privacy analysis}}
\label{sec:weekly segment}
\textcolor{black}{To prove that the synthetic data generated by our framework does not leak individual privacy, we perform experiments from three aspects~\cite{yuan2025learning}.}
\begin{packed_itemize}

\item \textcolor{black}{\textbf{Uniqueness testing~\cite{Yade}.}}
\textcolor{black}{This measure evaluates whether the generated data is completely identical to the original data. It highlights the extent to which the model directly generates copies instead of brand-new data.}

\textcolor{black}{To prove that the realistic generated mobility trajectory is not a simple copy of the real trajectory but a brand-new trajectory, we perform a uniqueness testing of it by comparing it with the real data. 
We randomly select generated trajectories and compare them with all the real trajectories from the training set. 
The two trajectories are aligned in the time dimension one by one and determine whether the locations at the 
corresponding time points are exactly the same. The overlapping ratio is defined as the ratio of the number of 
identical locations to the total trajectory length. Next, we choose The real trajectory that is most similar to the generated one is defined as the one with the highest overlapping ratio. We calculate the overlapping ratio distribution of all the generated trajectories with the most similar real trajectories mentioned before. 
The results can also be extended by considering more similar trajectories, e.g., the top-3 and top-5 most similar real trajectories.} 
\textcolor{black}{
As shown in Supplementary Figure\ref{fig:uniqueness study}, for the Smartphone datasets, more than 80\% of the generated mobility trajectories cannot find any real trajectories that have more than a 30\% overlapping ratio with them. For the Tencent dataset, more than 80\% of the generated mobility trajectories overlap with real trajectories with an overlapping ratio of less than 50\%. 
These results demonstrate that, while capturing mobility patterns, our framework indeed learns to generate brand-new and unique trajectories rather than simply copying.
}
\begin{figure}[t]
    \centering
    \subfigure[Smartphone]{
        \includegraphics[width=0.85\linewidth]{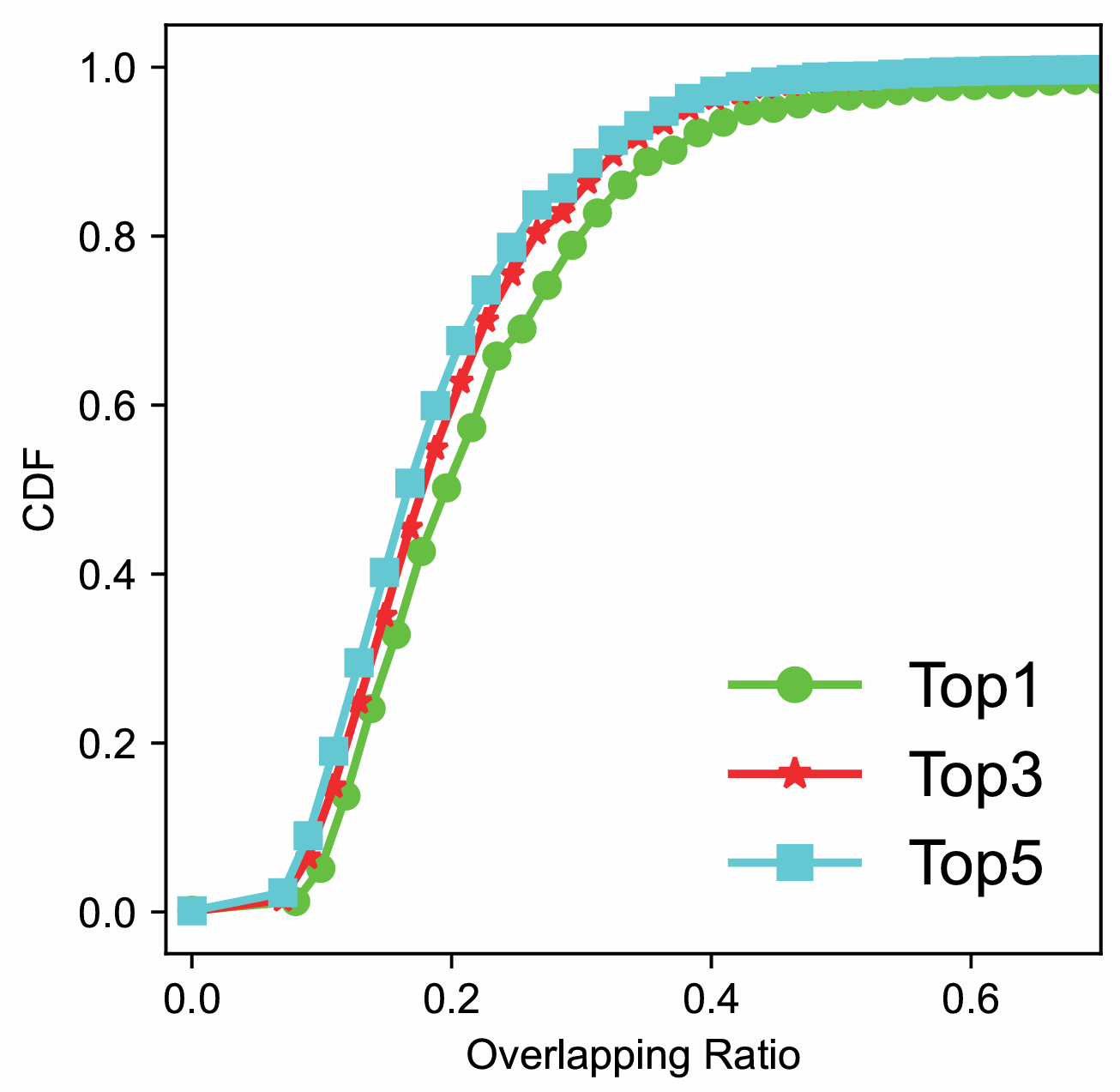} %
        \label{fig:smartphone_unique}
    }
    \hspace{0.05\linewidth} %
    \subfigure[Tencent]{
        \includegraphics[width=0.85\linewidth]{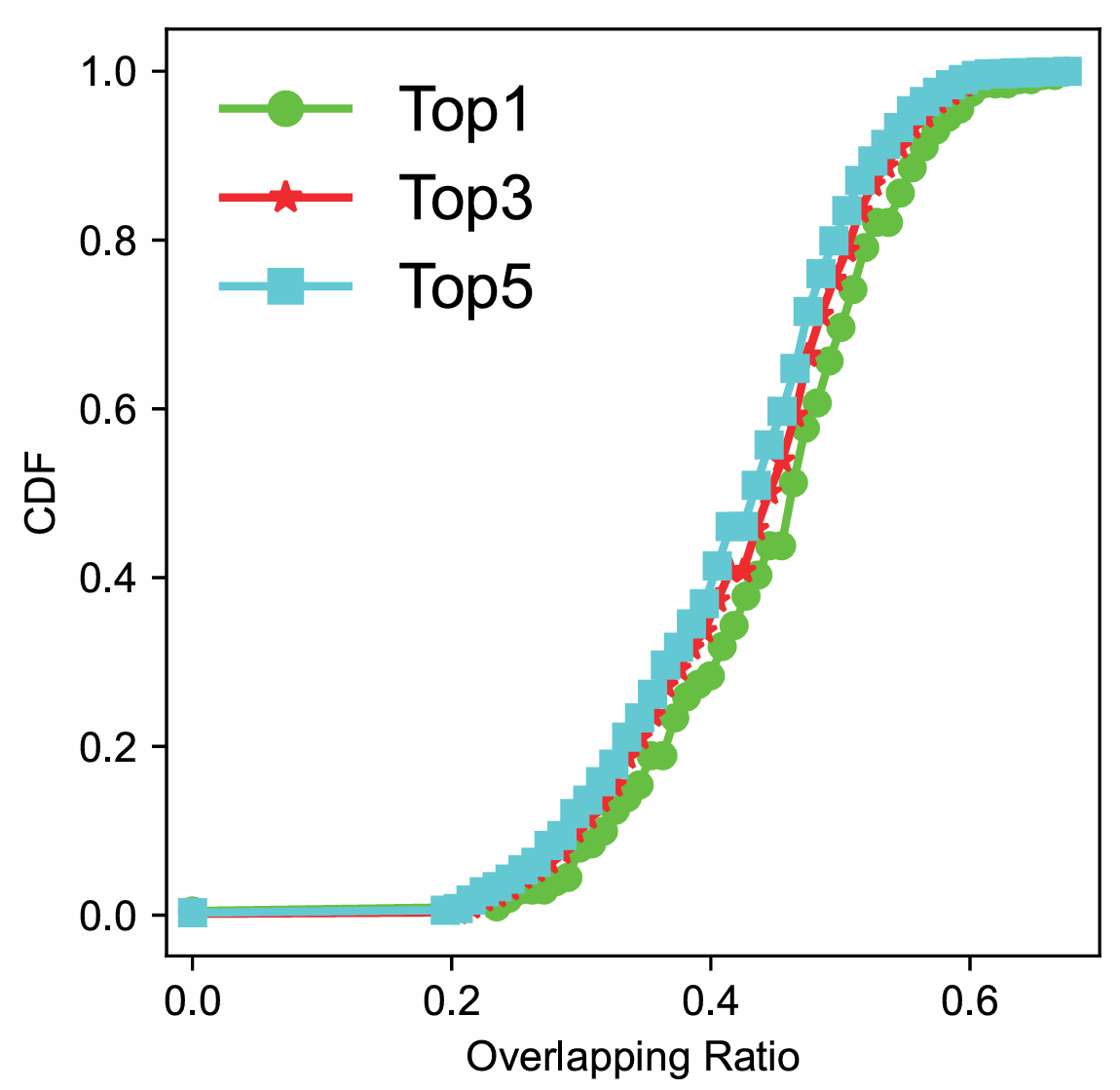} %
        \label{fig:tencent_unique}
    }
    \caption{Privacy evaluation in terms of uniqueness testing.}
    \label{fig:uniqueness study}
\end{figure}

\item \textcolor{black}{\textbf{Membership inference attack~\cite{7958568}.}}
\textcolor{black}{If the generated data does not reveal the identities of users from the original data, it should not be possible to use the generated data to reidentify users in the training set.For this purpose,we use the framework of membership inference attack~\cite{7958568}. Stronger privacy protection leads to a lower attack successrate.}

\textcolor{black}{Given a deep learning model and an individual record, the goal of the attack is to determine whether this record was included in the training set or not. 
We follow the attack settings as described in ~\cite{7958568}, where the attacker’s access to the deep learning model allows them to obtain the model’s output.
To improve the attack performance, we estimate individual information leakage using powerful machine learning models trained to predict whether an individual is in the training set. 
To control the impact of classification methods, we include four commonly used classification algorithms: Logistic Regression (LR), Support Vector Machine (SVM), 
K-Nearest Neighbors (KNN), and Random Forest (RF). 
The positive samples are those individuals in the training data, while the negative samples are not. 
The input feature is the overlapping ratio of multiple runs.
The evaluation metric is the success rate, defined as the percentage of successful trials in determining whether a sample is in the training set.
Stronger privacy protection leads to a lower success rate. 
As shown in Supplementary Figure\ref{fig:attack study}, on the Smartphone datasets the attack success rate is less than 0.55, and the Tencent dataset is less than 0.74. 
This result indicates that attackers can hardly infer whether individuals are in the training set based solely on the information of the generated urban mobility data. 
Thus, our framework demonstrates robustness against membership inference attacks.
}
\begin{figure}[t]
    \centering
    \subfigure[Smartphone]{
        \includegraphics[width=0.85\linewidth]{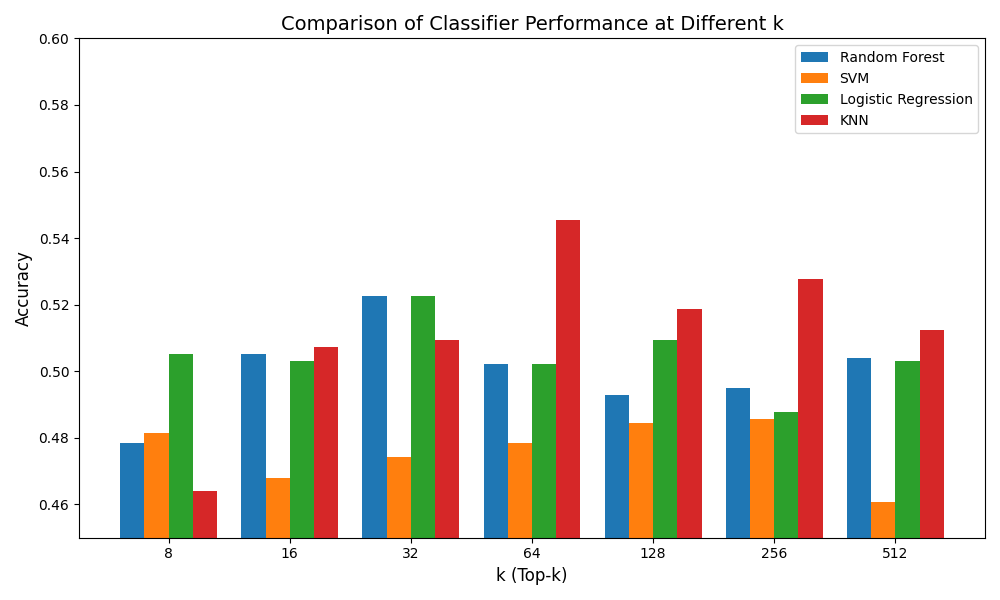} %
        \label{fig:smartphone_attack}
    }
    \hspace{0.05\linewidth} %
    \subfigure[Tencent]{
        \includegraphics[width=0.85\linewidth]{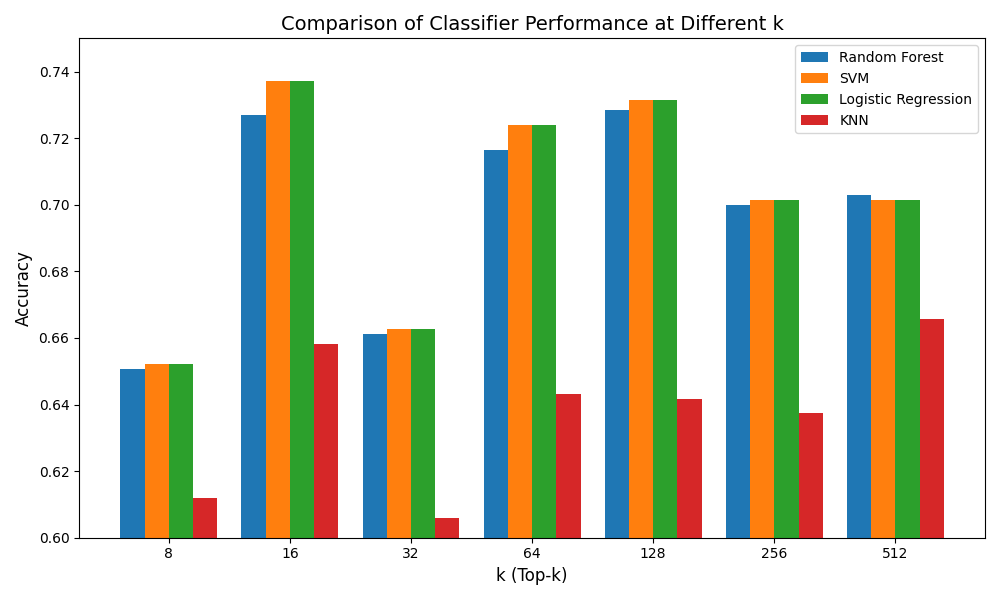} %
        \label{fig:tencent_attack}
    }
    \caption{Privacy evaluation in terms of Membership inference attack.}
    \label{fig:attack study}
\end{figure}

\item \textcolor{black}{\textbf{Differential privacy~\cite{10.1145/2976749.2978318}.}}
\textcolor{black}{A model is dierentially private if for any pair of training datasets and that differ in the  record of a single user, it holds that:$M(z;D) \leq e^{\epsilon} M(z;D') + \delta$ which means one can hardly distinguish whether any individual is included in the original dataset or not by looking at the output. It is a rigorous mathematical definition of privacy}

\textcolor{black}{For the output $z$, $M(z,D)$ denotes the probability distribution of $z$ with the data $D$ as the input. Smaller values of $\epsilon$ and $\delta$ provide stronger privacy guarantees. 
In our experiment, we examine the privacy budget of our proposed model from the perspective of changes in the overlapping ratio. Specifically, the overlapping ratio of each individual, under the conditions that this individual is included in the training set or not, is modeled by two Gaussian distributions, which are then regarded as $M(z,D)$ and $M(z,D')$ to calculate the privacy budget $\epsilon$. 
For each user, we compute $\epsilon$ using TensorFlow Privacy \cite{10.1145/2976749.2978318}. The cumulative distribution of $\epsilon$ is illustrated in Supplementary Figure\ref{fig:Differential study}. We observe that, without any additional privacy-preserving mechanism, when CDF is less than 0.9, our model achieves a maximum privacy budget of $\epsilon < 4$, which is typically considered a reasonable operating point for generative models. For example, Apple adopts a privacy budget of $\epsilon = 4.0$. The privacy budget can be further improved by incorporating DP-SGD or DP-GAN.
}

\begin{figure}[t]
    \centering
    \subfigure[Smartphone]{
        \includegraphics[width=0.85\linewidth]{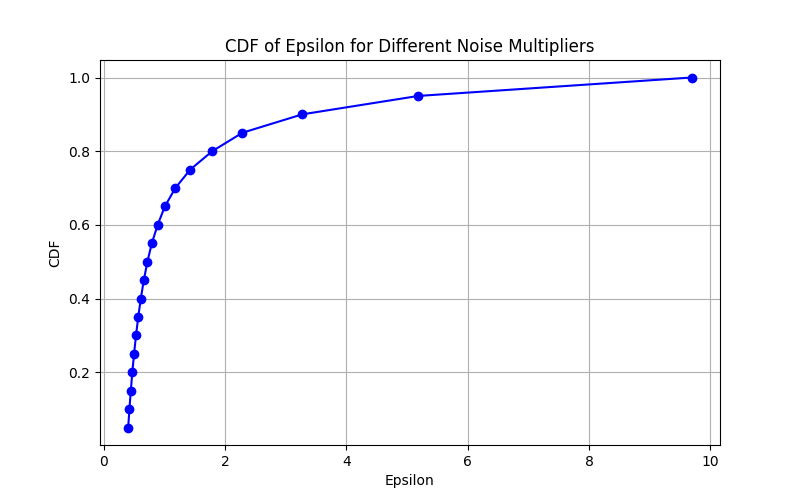} %
        \label{fig:smartphone_differential}
    }
    \hspace{0.05\linewidth} %
    \subfigure[Tencent]{
        \includegraphics[width=0.85\linewidth]{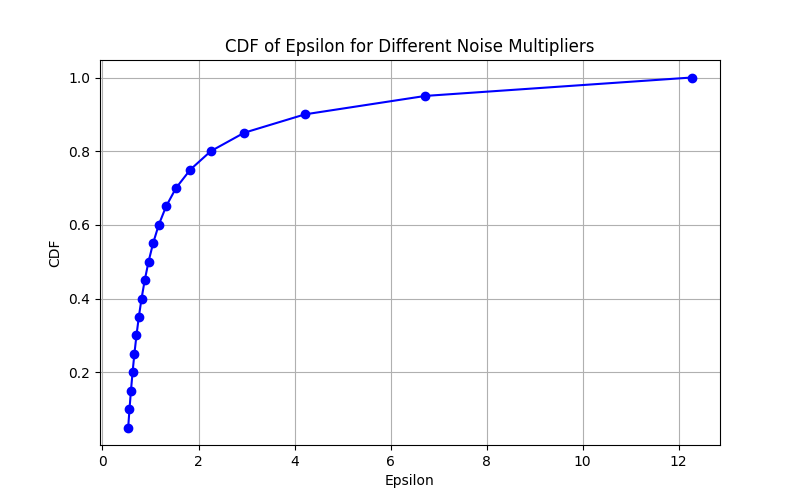} %
        \label{fig:tencent_differential}
    }
    \caption{Privacy evaluation in terms of Differential privacy.}
    \label{fig:Differential study}
\end{figure}

\end{packed_itemize}

\end{document}